\documentclass[runningheads]{llncs}

\usepackage{eccv}

\usepackage{eccvabbrv}

\usepackage{graphicx}
\usepackage{booktabs}
\usepackage{multirow}

\usepackage[accsupp]{axessibility}  

\usepackage{hyperref}

\usepackage{orcidlink}
\newcommand{\OURNAME}{DesignSense}

\begin{document}

\title{DesignSense: A Human Preference Dataset and Reward Modeling Framework for Graphic Layout Generation} 

\titlerunning{DesignSense}


\author{
Varun Gopal\inst{1}\thanks{Equal contribution. Contact: \email{msarkar@adobe.com}} \and
Rishabh Jain\inst{1}$^{\star}$ \and
Aradhya Mathur\inst{1} \and
Nikitha SR\inst{1} \and
Sohan Patnaik\inst{1} \and
Sudhir Yarram\inst{1} \and
Mayur Hemani\inst{1} \and
Balaji Krishnamurthy\inst{1} \and
Mausoom Sarkar\inst{1}
}

\authorrunning{V.~Gopal et al.}

\institute{
MDSR, Adobe
}



\maketitle

\begin{center}
    \includegraphics[width=\textwidth]{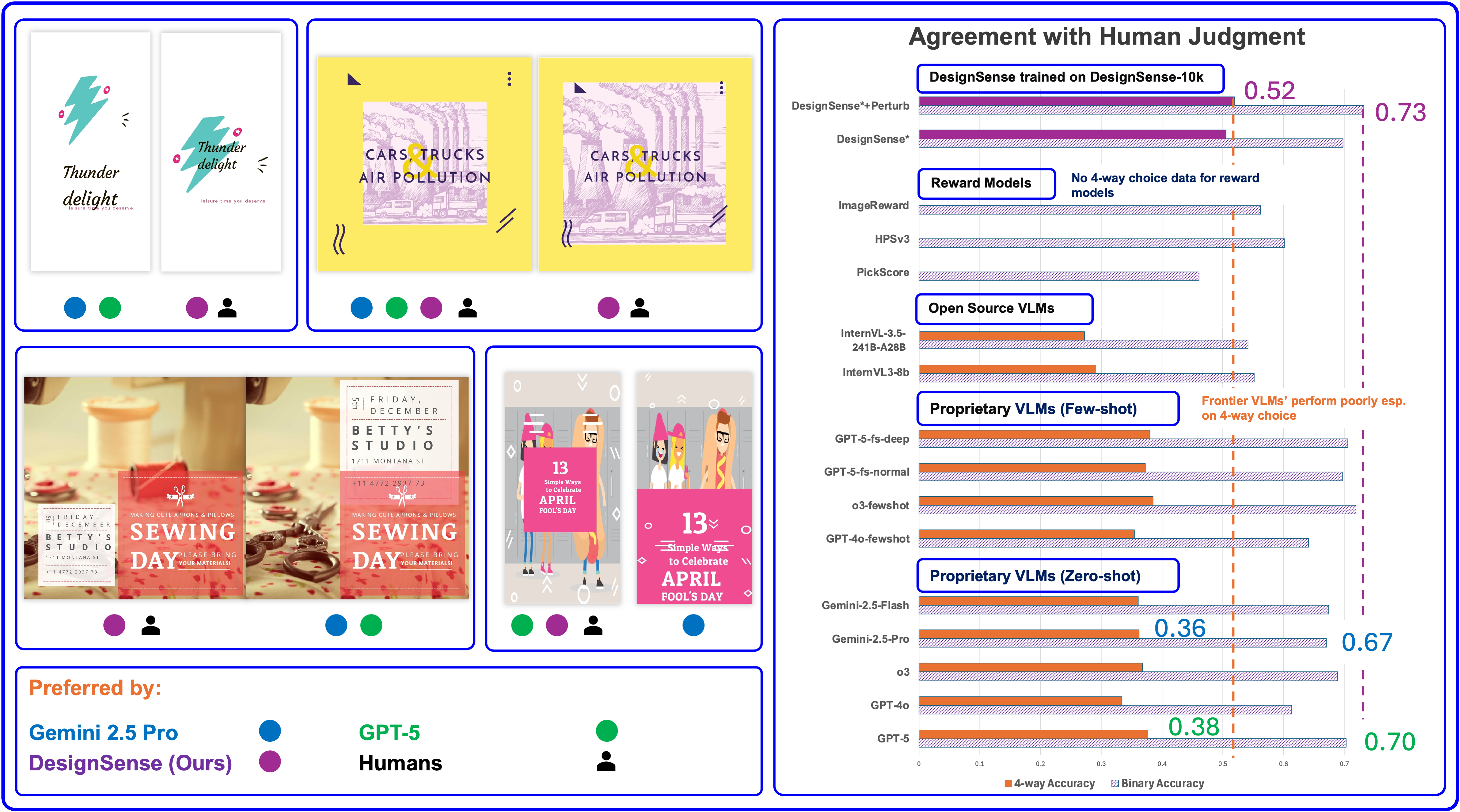}
    \captionsetup{type=figure}
    \captionof{figure}{
    Examples of predicted preference ordering by frontier vision-language models and our DesignSense model trained with the DesignSense-10k dataset.
    Each layout pair is evaluated through a 4-class annotation protocol: ``Left,'' ``Right,'' ``Both Good,'' and ``Both Bad.'' 
    The right panel summarizes model agreement with human judgments.
    (fs = few-shot; deep = deep thinking)
    }
\end{center}

\begin{abstract}
Graphic layouts serve as an important and engaging medium for visual communication across different channels. While recent layout generation models have demonstrated impressive capabilities, they frequently fail to align with nuanced human aesthetic judgment. Existing preference datasets and reward models trained on text-to-image generation do not generalize to layout evaluation, where the spatial arrangement of identical elements determines quality. To address this critical gap, we introduce DesignSense-10k, a large-scale dataset of 10,235 human-annotated preference pairs for graphic layout evaluation. We propose a five-stage curation pipeline that generates visually coherent layout transformations across diverse aspect ratios, using semantic grouping, layout prediction, filtering, clustering, and VLM-based refinement to produce high-quality comparison pairs. Human preferences are annotated using a 4-class scheme (“left,” “right,” “both good,” “both bad”) to capture subjective ambiguity. Leveraging this dataset, we train DesignSense, a vision-language model-based classifier that substantially outperforms existing open-source and proprietary models across comprehensive evaluation metrics (54.6\% improvement in Macro F1 over the strongest proprietary baseline). Our analysis shows that frontier VLMs remain unreliable overall and fail catastrophically on the full four-class task, underscoring the need for specialized, preference-aware models. Beyond the dataset, our reward model \textit{DesignSense} yields tangible downstream gains in layout generation. Using our judge during RL based training improves generator win rate by about 3\%, while inference-time scaling, which involves generating multiple candidates and selecting the best one, provides a 3.6\% improvement. These results highlight the practical impact of specialized, layout-aware preference modeling on real-world layout generation quality.
\end{abstract}

\section{Introduction}
\label{sec:intro}

 Graphical layouts serve to communicate information in an engaging manner across diverse applications including advertising, poster design, entertainment, and digital art. Recent advances in deep learning based models ~\cite{layoutdm,patnaik2025aesthetiq,yu2024layoutdetr} have made significant progress in speeding up the process of producing aesthetically pleasing layouts from visual and textual elements. In this work, we address the problem of aligning these models with human aesthetic preferences.
The alignment of image generation models with human visual preferences has been the focus of several studies that propose datasets (ImageRewardDB \cite{imagereward}, HPD \cite{hpsv2}, Pick-a-pic \cite{pickapic}) and models (HPS \cite{hpsv2, hpsv3}, ImageReward \cite{imagereward}, PickScore \cite{pickapic}, and HP Score \cite{pick_high}) that work as proxies for human evaluations. These models learn from pairwise comparisons of images generated by text-to-image diffusion models spanning diverse visual domains. However, our empirical evaluation (Table~\ref{tab:annotation_results}) reveals that these models fail to capture human preferences for graphic layouts. We attribute this to a fundamental distributional mismatch. Layout preference depends on spatial relationships, compositional balance, and hierarchical organization, all of which reflect design intent rather than the photorealistic content that dominates the training data of existing preference models. Since these models are trained primarily on natural images, they lack exposure to the structural cues that guide layout aesthetics. This gap is further exacerbated by the absence of large-scale datasets capturing human preferences for layout design.
To address this limitation, we introduce \OURNAME, a large-scale dataset of human layout preferences. While inspired by pairwise preference datasets used in text-to-image generation, our formulation is richer: annotators evaluate each layout pair using four labels-left, right, both good, and both bad, capturing not only directional preference but also aesthetic ambiguity and cases where both designs succeed or fail. Such comparisons effectively capture preferences across key design dimensions, including compositional balance, alignment, overlap, and visual hierarchy. Constructing these pairs, however, requires generating layout variants that differ meaningfully along these factors.

To enable this, we propose a novel five-stage data transformation pipeline built on the Crello dataset \cite{Yamaguchi2021CanvasVAELT} to generate a large set of high-quality comparison pairs by leveraging state-of-the-art layout synthesis model AesthetiQ \cite{patnaik2025aesthetiq}.  The pipeline consists of five stages:  i) Grouping related elements using semantic clustering to reduce the number of inputs to the layout model; ii) Generating candidate layouts using the layout model ; iii) Filtering out low-quality layout candidates; iv) Ensuring diversity through clustering followed by sampling; and v) Fine-tuning layouts to eliminate overlaps and improving object alignments. This pipeline enables efficient generation of preference pairs across various aspect ratios while maintaining high aesthetic standards suitable for human annotation.
Building on \OURNAME, we introduce a vision-language-model-based judge titled \textit{DesignSense} that significantly outperforms both open-source and proprietary baselines across comprehensive evaluation metrics, achieving a 54.6\% improvement in Macro F1 over the strongest proprietary model. Moreover, \textit{DesignSense} enables meaningful downstream gains in improvement layout generation. When incorporated into reinforcement learning frameworks as a stronger preference judge, it improves the generation abilities of the model such as AesthetiQ \cite{patnaik2025aesthetiq} significantly.  In addition, the reward model supports effective inference-time scaling: by generating multiple layout candidates and ranking them through \textit{DesignSense}, the system can reliably select the highest-quality output. These applications illustrate the practical value of specialized, layout-specific preference learning for achieving human-aligned visual design generation.
Our main contributions are:
\begin{itemize}
    \item \OURNAME, the first large-scale dataset of 10,235 human-annotated preference pairs for layout quality assessment, featuring images with diverse aspect ratios and consistent element-sets with varied spatial arrangements.
    \item A novel five-stage data curation pipeline for high-quality layout generation across diverse aspect ratios, incorporating grouping, prediction, refinement, filtering, and clustering stages.
    \item Comprehensive evaluation of existing foundation models alongside a baseline model \textit{DesignSense} trained on the proposed dataset, demonstrating its effectiveness over them.
    \item Additionally, we show that \textit{DesignSense} improves win rate of layout generation models like AesthetiQ \cite{patnaik2025aesthetiq} by about 3\%, and independently enables effective inference-time scaling that provides a 3.6\% improvement, leading to more human-aligned layouts.
\end{itemize}

\section{Related Work}
\label{sec:related_work}

\subsection{Layout Generation Models and Benchmarks}
\label{ssec:layout_gen}

Layout generation models have evolved significantly over the past decade. Early works employed classical energy-based methods focused on design principles~\cite{o2014learning}, and GAN-based content-aware generation ~\cite{zheng2019content}.
Subsequent work has focused on integrating user and design constraints~\cite{Lee2019NeuralDN, Kikuchi2021ConstrainedGL}. Recent approaches include diffusion-based models, which use latent diffusion~\cite{pLay} or discrete diffusion~\cite{layoutdm, ldgm} frameworks, and transformer backbones~\cite{Zhang2023LayoutDiffusionIG, lace} to improve synthesis. 

On the other hand, \textit{Large language models} have been explored for layout generation by treating layouts as structured data formats. For example, LayoutNUWA~\cite{tang2024layoutnuwa} fine-tunes LLaMA~\cite{Touvron2023LLaMAOA} and CodeLLaMA~\cite{Rozire2023CodeLlama} for content-agnostic generation. PosterLlama~\cite{posterllama} reformats inputs elements into HTML. Other methods like CanvasVAE~\cite{Yamaguchi2021CanvasVAELT} , FlexDM~\cite{flexdm} , and ICVT~\cite{icvt} integrate multimodal information, using transformers to incorporate images and text. The state-of-the-art AesthetiQ~\cite{patnaik2025aesthetiq} leverages a VLM backbone and uses ViLA\cite{10657989} for preference-aware alignment and evaluation.

Despite these advances, existing models fail to align with human preferences, constrained by the lack of diverse, preference-annotated layout data. Current benchmarks like LayoutBench~\cite{layoutbench}, SciPostLayout~\cite{tanaka2024scipostlayout}, and the CGL-Dataset~\cite{cgldataset} are valuable, but focus on spatial control or object relationships rather than aesthetics. To our knowledge, no large-scale, human-preference dataset for layout aesthetics exists.

\subsection{Human Visual Preference Datasets and Models}
\label{ssec:human_pref}

\textit{Image generation preference learning} has gained significant attention for aligning text-to-image models with human preferences. Metrics such as FID~\cite{fid}, Inception Score~\cite{inceptionscore}, CLIPScore~\cite{hessel2021clipscore} are used to assess the quality of image generations. Several datasets such as HPDv1~\cite{hpdv1}, ImageReward~\cite{imagereward}, Pick-a-Pick~\cite{pickapic} have comparisons of the images generated by various diffusion models. HPDv2~\cite{hpsv2}, HPDv3~\cite{hpsv3} involve human annotated pairs of images of a wider range of models. For Instance, HPDv3 consists of 1.08 million text-image pairs and 1.17 million annotated pairwise comparisons, achieving superior alignment with human judgment.

Models to learn human preferences, such as PickScore~\cite{pickapic}, HPS~\cite{hps}, HPSv2~\cite{hpsv2}, and MPS~\cite{mps}, fine-tune CLIP-based~\cite{clip} architectures, while ImageReward~\cite{imagereward} utilizes a BLIP~\cite{blip} encoder to train its reward model. HPSv3~\cite{hpsv3} achieves strong alignment by leveraging a Vision Language Model (VLM) backbone trained on HPDv3. This field has also extended to video generation with datasets like SafeSora~\cite{safesora} and models like VisionReward~\cite{visionreward}.


However, all these preference datasets and models focus exclusively on text-to-image generation, where paired images differ in their constituent elements and visual content. The evaluation criteria emphasize image fidelity, prompt alignment, and aesthetic quality of photorealistic images. These models fail to generalize to layout comparison tasks where the same elements appear with different spatial arrangements, representing a fundamentally different distribution from typical image generation outputs.


\subsection{Multimodal Reward Models}
\label{ssec:multimodal_reward}

Recent efforts in multimodal preference alignment have focused on enhancing VLM outputs through various alignment strategies. This includes developing general-purpose multimodal reward models \cite{ma2024mpr, mgeneralr}, providing supervised signals \cite{wang2025skywork} for MPO \cite{mpo}, using RLHF with additional context \cite{llavarlhf}, generating automatic preference labels \cite{ixc}, and introducing process-level reward signals \cite{wang2025visualprm}. Other works have focused on developing robust RLHF pipelines specifically for vision-language models \cite{yu2024rlhfv}.




However, existing multimodal reward models are trained on VQA and reasoning tasks with primarily textual outputs , lacking the specialized understanding for layout aesthetics where factors like spatial arrangement, visual hierarchy, and balance are critical. This gap motivates our dataset, the first large-scale resource for training layout-specific reward models aligned with human aesthetic judgment
\section{Approach}

In this section, we present our pipeline for generating high-quality, diverse, aesthetically pleasing graphic layout pairs with different aspect ratios that can be used to train layout preference models, and the details of the construction process of the \OURNAME dataset. 

\begin{figure*}[t!]
    \centering
    
    \includegraphics[width=\linewidth]{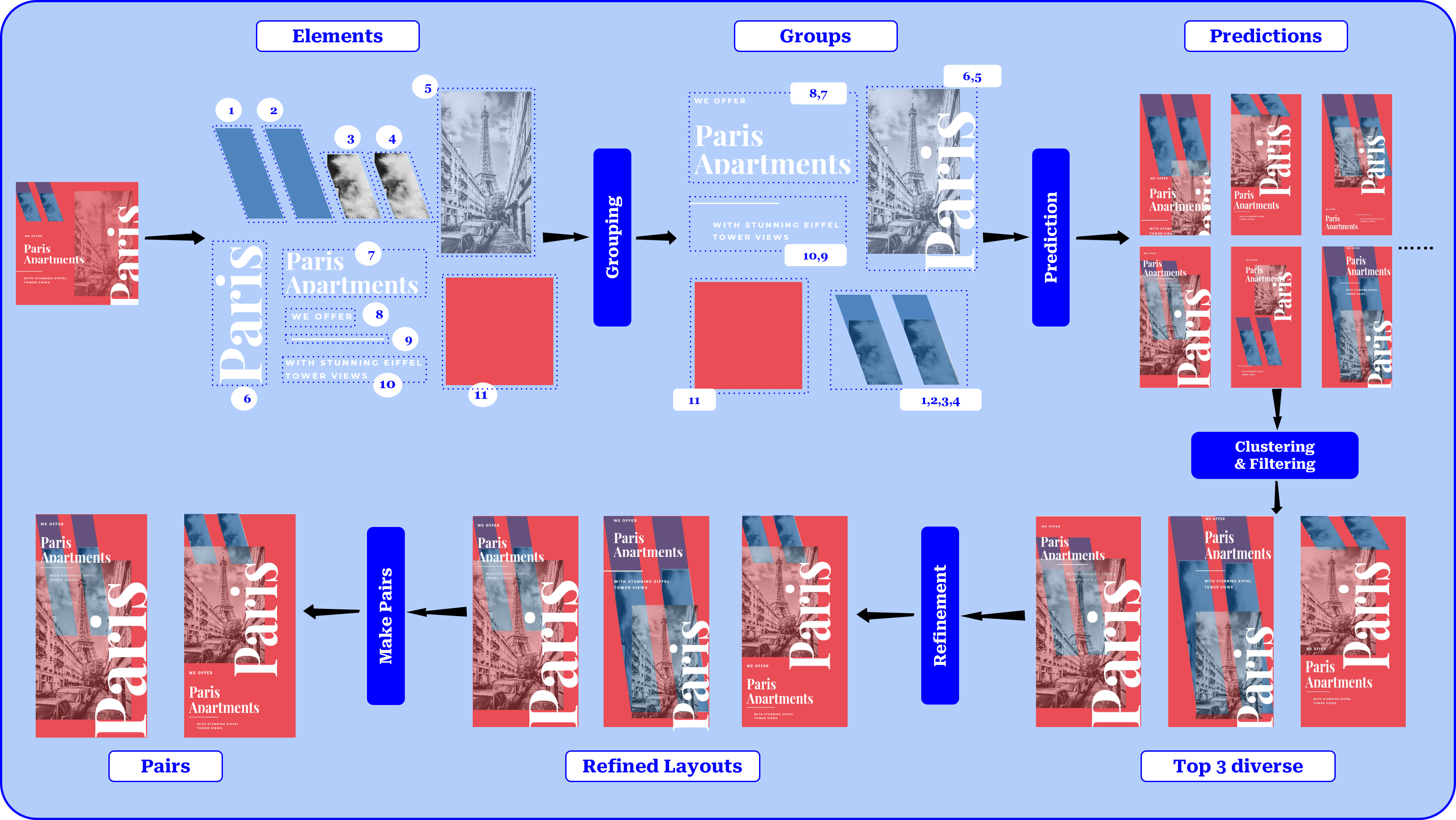}
    \caption{
Overview of the DesignSense data curation pipeline, illustrated in five main steps:\textbf{Step 1: Grouping} - Elements from the original layout are grouped based on semantic and spatial relationships to reduce structural complexity and preserve design intent.\textbf{Step 2: Prediction} - Grouped elements are fed into a layout prediction model to generate multiple candidate relayouts under diverse aspect ratio conditions. \textbf{Step 3: Clustering \& Filtering} - Generated layouts are clustered to maximize output diversity and filtered to retain only high-quality candidates, selecting the top three most distinct layouts for each setting.
\textbf{Step 4: Refinement} - Selected layouts are further improved using a refinement module which optimizes element positions, resolves overlaps, and enhances overall visual alignment. This end-to-end process enables the construction of a large-scale, diverse, and preference-annotated graphic layout dataset.
}

    \label{fig:pipeline}
\end{figure*}

\subsection{Layout Generation Pipeline}
Our framework for layout pair generation begins with the Crello dataset~\cite{Yamaguchi2021CanvasVAELT}, which contains approximately 19.3k layouts consisting of text and image elements along with their corresponding bounding boxes. These layouts are transformed across diverse aspect ratios through a systematic five-stage pipeline: 
(i) \textbf{Grouping} --- semantic clustering related elements to reduce prediction complexity; 
(ii) \textbf{Prediction} --- generating candidate layouts using a state-of-the-art layout prediction model (AesthetiQ~\cite{patnaik2025aesthetiq}) referred henceforth as the Layout model; 
(iii) \textbf{Filtering} --- removing low-quality predictions; 
(iv) \textbf{Clustering} --- to select diverse layouts for layout pairs; and 
(v) \textbf{Refinement} --- fine-tuning layouts to eliminate overlaps and improve alignment. Figure \ref{fig:pipeline} illustrates the process with an example. We describe each of these stages in detail below.

\paragraph{Grouping} The Layout model predicts position tokens for individual design elements on a discrete canvas. We observe that the quality of predictions from the Layout model improves for fewer elements. Accordingly, we semantically group related components, such as a placeholder and its associated text or a date and time of an event. 
Formally, grouping is defined as a function $f_{\text{group}}: \mathcal{E} \rightarrow \mathcal{G}$, where $\mathcal{E}$ denotes all elements in a layout and $\mathcal{G} = \{G_1, G_2, \ldots, G_k\}$ represents $k$ groups such that:
\begin{align}
    \bigcup_{i=1}^{k} G_i &= \mathcal{E}, \\
    G_i \cap G_j &= \emptyset \quad \forall i \neq j, \\
    G_i &\neq \emptyset \quad \forall i.
\end{align}
This grouping stage reduces combinatorial complexity for the layout prediction task and improves contextual coherence during generation.We use the GPT-4o~\cite{openai2024gpt4o} model to identify and group elements that should move together based on semantic relationships, spatial proximity, and visual hierarchy. 
To evaluate grouping quality, we use the Adjusted Rand Index (ARI)\footnote{ARI is used to measure similarity between two clusterings.}. We manually annotate 43 layouts, compare different grouping models and fewshot settings, and find that GPT-4o~\cite{openai2024gpt4o} with one-shot setting achieves the highest score of $\sim$0.69, indicating its alignment with human judgment. Implementation and evaluation details are provided in the Appendix.

\paragraph{Prediction} We retrain the Layout generation model AesthetiQ \cite{patnaik2025aesthetiq} to operate at the \textit{group level} rather than the individual element level using the dataset from the grouping stage, enabling more coherent and semantically consistent layout generation. The retrained Layout model $f_{\text{pred}}$ takes as input the grouped element images $\{\mathbf{G}_1, \mathbf{G}_2, \ldots, \mathbf{G}_k\}$ and the target aspect ratio $r_{\text{target}}$, and outputs bounding box predictions $\{\mathbf{b}_1, \mathbf{b}_2, \ldots, \mathbf{b}_k\}$:
\begin{equation}
    \{\mathbf{b}_i\}_{i=1}^{k} = f_{\text{pred}}(\{\mathbf{G}_i\}_{i=1}^{k}, r_{\text{target}} \mid \theta),
\end{equation}
where $\theta$ denotes the model parameters. This adaptation allows the Layout model to reason over grouped visual units, resulting in improved structural alignment and overall layout quality. To ensure diversity among the generated variants, we sample ten candidate layouts for each input element set using temperature-based decoding, then apply filtering and clustering to select the most suitable variants.

\paragraph{Filtering and Clustering} 
The filtering step removes samples with layout issues like overlaps and overflows, utilizing the GPT-4o~\cite{openai2024gpt4o} model as an intermediate judge model. This eliminates clearly suboptimal layouts (easy negatives) from the dataset.


To promote greater diversity among the retained layouts, we perform IoU-based clustering to group similar designs and select representative examples from each cluster. Two layouts are considered similar if their corresponding elements exhibit high positional overlap, and clusters are merged when they share a high average pairwise similarity across their layouts. From each cluster, the layout with the highest mean similarity to all other members is selected as the representative. This process produces a compact yet diverse subset of high-quality layouts, well-suited for human evaluation (detailed algorithm present in Appendix).

\paragraph{Refinement} Most layouts produced in the prediction stage are visually coherent. However, minor imperfections such as subtle overlaps, uneven spacing, or slight misalignments may persist. The \textit{refinement stage} uses the OpenAI o3 model \cite{openai2025o3} to adjust group-positions while preserving the intended design. Each rendered layout and the corresponding metadata are given as input to the model, to obtain refined bounding boxes as output $\{\mathbf{b}_i'\}_{i=1}^{k} = f_{\text{refine}}(\{\mathbf{b}_i\}_{i=1}^{k}, \{\mathbf{G}_i\}_{i=1}^{k})$, minimizing overlaps and improving alignment. The exact prompt settings and evaluation details are included in the Appendix. The refinement process ensures that the resulting layouts are sufficiently aesthetically polished before human annotation.
To assess the effectiveness of the refinement stage, we curate approximately 200 pairs of predicted layouts and their corresponding refined versions, and conduct human preference evaluations. We measure the ratio of the number of pairs in which the refined layout is preferred to the times when the original is favored.
This \textit{human preference ratio} (HPR) is used for choosing the best few-shot prompting setting from among several options.
The detailed analysis is provided in the appendix. 

These five stages of data refinement are used to construct the \OURNAME dataset. Next, we detail the design choices in the process.

\begin{table*}[t]
\centering
\begin{tabular}{lccccc}
\toprule
Model & Accuracy$\uparrow$ & Binary Acc$\uparrow$ & Cohen's $\kappa$$\uparrow$ & Macro F1$\uparrow$ & Weighted F1$\uparrow$ \\
\midrule

\multicolumn{6}{l}{\textit{Open-Source Models}} \\
InternVL3-8B~\cite{zhu2025internvl3exploringadvancedtraining} & 0.291 & 0.553 & 0.032 & 0.208 & 0.207 \\
InternVL-3.5-241B-A28B~\cite{wang2025internvl3_5}  & 0.273 & 0.543 & 0.027 & 0.21 & 0.197 \\
ImageReward~\cite{imagereward} & - & 0.562 & - & - & - \\
HPSv3~\cite{hpsv3} & - & 0.602 & - & - & - \\
PickScore~\cite{pickapic} & - & 0.461 & - & - & - \\
\midrule 

\multicolumn{6}{l}{\textit{Proprietary Models}} \\
GPT-4o~\cite{openai2024gpt4o} & 0.334 & 0.613 & 0.100 & 0.275 & 0.249 \\
OpenAI-o3~\cite{openai2025o3} & 0.368 & 0.689 & 0.142 & 0.278 & 0.267 \\
GPT-5-fewshot-normal~\cite{openai2025gpt5} & 0.373 & 0.697 & 0.149 & 0.273 & 0.269 \\
GPT-5~\cite{openai2025gpt5} & 0.377 & 0.703 & 0.156 & 0.270 & 0.270 \\
GPT-4o-fewshot~\cite{openai2024gpt4o} & 0.354 & 0.641 & 0.125 & 0.292 & 0.271 \\
Gemini-2.5-flash~\cite{google2025gemini2.5flash} & 0.361 & 0.674 & 0.131 & 0.289 & 0.274 \\
GPT-5-fewshot-deepthink~\cite{openai2025gpt5} & 0.380 & 0.705 & 0.159 & 0.295 & 0.276 \\
Gemini-2.5-pro~\cite{deepmind2025gemini2.5pro} & 0.362 & 0.670 & 0.132 & 0.295 & 0.278 \\
OpenAI-o3-fewshot~\cite{openai2025o3} & 0.385 & \underline{0.719} & 0.164 & 0.283 & 0.279 \\
\midrule 
\midrule

DesignSense (w/o Perturb) & \underline{0.505} & 0.698 & \underline{0.239} & \underline{0.406} & \underline{0.484} \\
\textbf{DesignSense (Ours)} & \textbf{0.519} & \textbf{0.732} & \textbf{0.292} & \textbf{0.456} & \textbf{0.520} \\
\bottomrule
\end{tabular}
\caption{
Quantitative results for the 4-class layout preference task, comparing our DesignSense model against open-source and proprietary baselines. Our method achieves state-of-the-art performance, significantly outperforming all other models across all reported metrics. The best scores are in \textbf{bold}, and the second-best are \underline{underlined}.
}
\label{tab:annotation_results}
\end{table*} 

\subsection{\OURNAME Dataset Construction}

\paragraph{Choice of Variants} The Layout model takes the canvas size as input, in addition to the elements to be arranged. This allows us to produce variants in different aspect ratio settings. The original dataset has about 19.3k layout samples. In the Prediction stage of the pipeline, we choose two different aspect ratio settings besides the original size (for a total of 3 settings):

\begin{itemize}
    \item \textbf{Stretching-2x}: the longest side is scaled by a factor of 2, creating an elongated aspect ratio.
    \item \textbf{Inverse-Ratio}: the dimensions of the longer and shorter sides are swapped, producing a transposed canvas orientation.
\end{itemize}

\paragraph{Sampling}
A total of approximately 150,000 layout images (three aspect ratio settings, each with about 50,000 samples) are generated across the pipeline stages. To promote dataset diversity and ensure meaningful human evaluation, we employ the DINO v3 model~\cite{simeoni2025dinov3} to extract features for each layout. Pairwise similarity scores are computed based on these representations, and we select the 10k most diverse pairs (i.e., those with the lowest similarity scores). Specifically, we sample 4k pairs each from the Stretching-2x and Inverse-Ratio variants, which exhibit the greatest layout variation due to significant aspect ratio transformations. The remaining 2000 pairs are drawn from the subset with the original aspect ratios, where layouts are typically less varied as their element composition was originally tailored to the native aspect ratio. This emphasis on aspect ratio challenges the Layout model to adapt element scaling and positioning to new design constraints.

\paragraph{Human Annotation}
The final dataset of $10$k preference pairs is presented to human annotators to capture their aesthetic judgment. The pairs are presented side-by-side to each annotator and they are asked to indicate their preference by choosing one of four choices: ``$left$", ``$right$", ``$both\_good$", ``$both\_bad$", where the first two choices allow direct selection based on better quality, and the last two choices allow subjective assessment in ambiguous cases.


\begin{figure*}[t]
    \centering
    
    \includegraphics[width=0.95\linewidth]{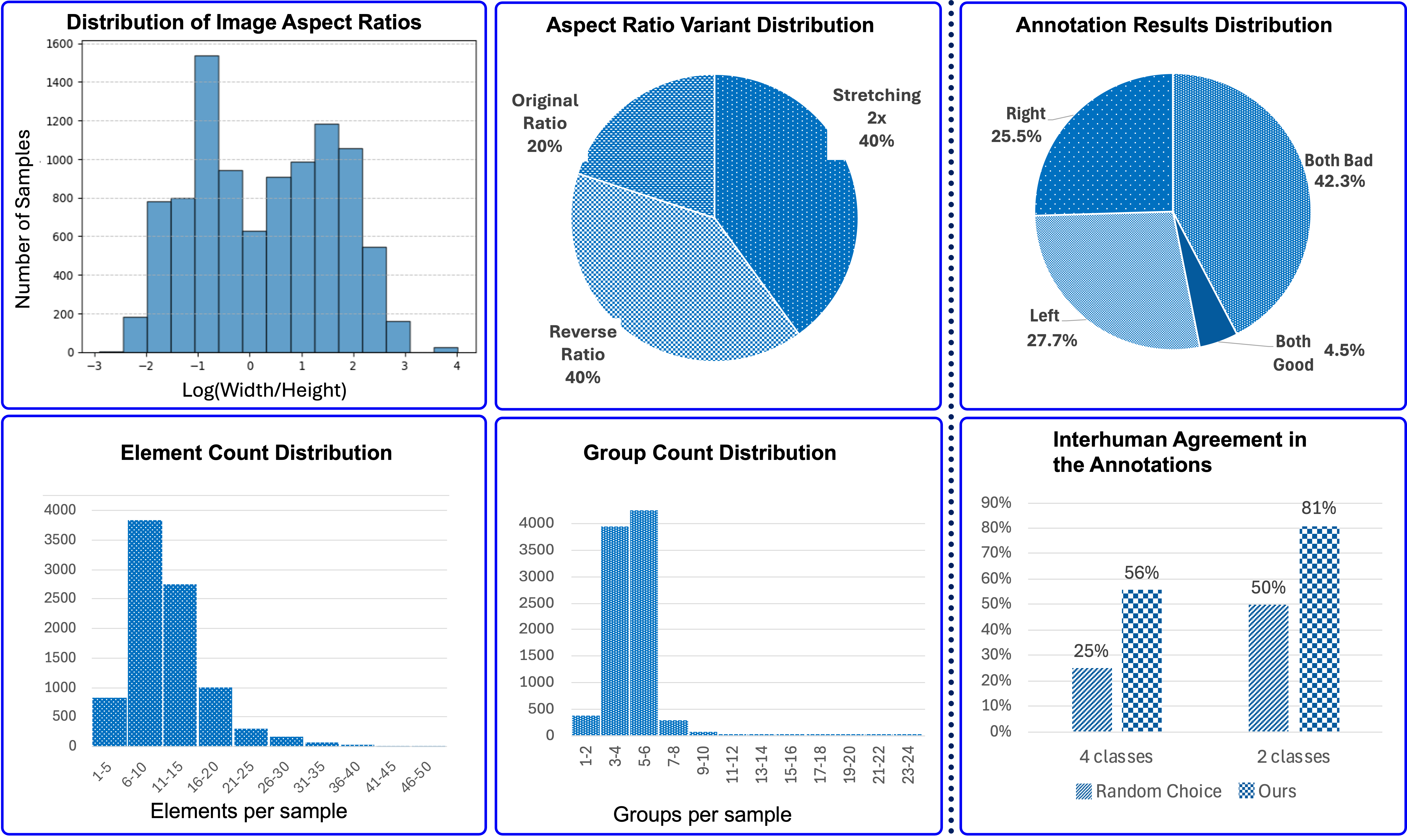}
\caption{
Dataset statistics and annotation analysis for DesignSense. Top left: Distribution of image aspect ratios (log$_2$(width/height)) illustrating the diversity of layouts. Top right: Pie charts showing the distribution of relayout settings (“Stretching\_2x,” “Reverse Ratio,” “Original Ratio”) and annotation result classes (“Both Bad,” “Both Good,” “Left,” “Right”). Bottom left: Histogram of number of groups per sample, and number of elements per sample, highlighting compositional complexity. Bottom right: Interhuman agreement illustrating substantially higher consistency among our annotators compared to random choice, for both 4-class and 2-class settings.
}

    \label{fig:dataset_stats}
\end{figure*}

\paragraph{Annotation Quality Evaluation:}
We use the Amazon Mechanical Turk (AMT) platform to obtain annotations for the preference pairs. To mitigate annotation inconsistency, we perform an initial test annotation exercise with about 100 samples and 5 annotators, collecting 500 annotations. The 4-way (``left'', ``right'', ``both\_good'', ``both\_bad'') and binary (``left''\/``right'') inter-human agreement for the test was 56\% and 81\% respectively, indicating that the results are adequately consistent. Approximately 80\% of the dataset is annotated through AMT, and the rest was annotated by the authors for high-quality comparison pairs.



\paragraph{Dataset Statistics} 
The final curated dataset comprises 10,235 paired graphic layouts with associated human preference annotations. Of these, 8,735 pairs are used for training the DesignSense classifier, 500 pairs form the validation set, and 1,000 pairs are held-out as the test split. Figure~\ref{fig:dataset_stats} charts the distribution of the dataset. The samples have diverse sizes and aspect ratios, which promotes robustness to compositional variability and ensures that our judge model generalizes effectively across different layout formats. Notably, a significant number of pairs were still labeled as "\textit{both bad}" even after meticulous data preparation, underscoring the need for a strong judge to provide reliable reinforcement signals for improving layout generation models.




\section{Experiments}
The value of the \OURNAME dataset is evaluated by training a classifier model which is based on a pre-trained InternVL3-8B model \cite{zhu2025internvl3exploringadvancedtraining}. In this section, describe its training details and its evaluation with respect to ground-truth human preferences.

\subsection{Training the DesignSense Classifier}
To model human aesthetic and functional preferences in layout design, we train a vision-language model (VLM)-based \textbf{DesignSense} built upon the InternVL3-8B backbone~\cite{zhu2025internvl3exploringadvancedtraining}. Unlike vision-based classifiers that rely solely on visual similarity, the VLM-based model reasons about layout composition, balance, and design intent through natural language understanding. The model is fine-tuned on the training split of the \OURNAME dataset. Each sample in the dataset consists of a pair of layouts and one of four corresponding human preference label \{\textit{left is better}, \textit{right is better}, \textit{both\_good}, and \textit{both\_bad}\}. During training, the model receives both rendered layouts and textual metadata, and is prompted to pay attention to the visual appeal, spacing and alignment, element clarity, consistency with design principles and patterns. A tie-breaking rule is included in the prompt to prioritize the absence of visual defects. The exact prompt is included in the Appendix.




\paragraph{Dataset augmentation:} To enhance generalization and robustness, we further augment the dataset using a \textit{perturbation strategy}. Specifically, we introduce controlled degradations into ground-truth templates by randomly selecting $~70\%$ of the elements and perturbing their bounding boxes. For each selected element, we applied one of two random perturbations: either a position offset (a random bidirectional shift of 20-50\% relative to the element's size), or a scale perturbation (a random scaling factor between 0.8x and 1.2x) for the element's width or its height (both equally likely). The templates are then rendered with these modified attributes. These perturbed versions are automatically labeled as negative samples (\textit{bad layouts}), providing the model with additional contrastive supervision to distinguish design flaws such as misalignment, uneven spacing, or visual imbalance. 


\subsection{Benchmarking}
We benchmark our DesignSense model against a comprehensive suite of baselines and report relevant evaluation metrics for each (see Table \ref{tab:annotation_results}). The model outperforms all baselines (even the best reasoning models), establishing the need for specialized fine-tuning for tasks like layout preference prediction.

\paragraph{Baseline models:}
The baselines include open-source and proprietary vision-language models (VLMs) . The open-source category includes InternVL3.5-241B as a strong general-purpose baseline. We also compare the model's accuracy with established preference and aesthetic reward models, namely ImageReward~\cite{imagereward}, HPSv3~\cite{hpsv3}, and PickScore~\cite{pickapic}. The proprietary category features state-of-the-art VLMs from OpenAI (e.g., GPT-4o~\cite{openai2024gpt4o}, GPT-5~\cite{openai2025gpt5}, OpenAI-o3~\cite{openai2025o3}), Google (e.g., Gemini-2.5-Pro~\cite{deepmind2025gemini2.5pro}). We evaluate these proprietary models in both zero-shot (e.g., GPT-4o) and few-shot (e.g., GPT-4o-fewshot, OpenAI-o3-fewshot) configurations. For a fair comparison, we include the instruction examples shared with the human annotators in the prompts for the few-shot configurations. This also tests the candidate models' ability to learn the nuanced layout preference task from the same limited context provided to human experts rather than relying solely on its pre-trained knowledge.
\paragraph{Evaluation Tasks:} The models are evaluated on two tasks - the two-way selection task for each layout pair, and a 4-class layout preference selection task that additionally allows rating the two layouts as either \textit{both\_good} or \textit{both\_bad}. The open-source reward models (PickScore, ImageReward, and HPSv3) are primarily designed to output a scalar reward or binary choice, so their evaluation is focused on the binary selection task.

\paragraph{Evaluation Metrics:}  To evaluate model performance on our 4-class layout preference task, we report a comprehensive set of standard and task-specific metrics. Due to the significant class imbalance inherent in preference data (where "both are good" or "both are bad" may be far more or less common than a specific choice), we report Macro F1~\cite{pedregosa2011scikit} and Weighted F1~\cite{pedregosa2011scikit} scores. The Macro F1 score is particularly important as it computes the F1 score for each class independently and averages them, treating each class equally regardless of its sample size. The Weighted F1 score, in contrast, weights the F1 score of each class by its support. We also report overall Accuracy and Cohen's $\kappa$~\cite{cohen1960coefficient}, which measures inter-annotator agreement (in this case, model vs. human) while correcting for agreement that could occur by chance. Furthermore, to specifically assess a model's ability to make a correct choice when a clear preference is expressed, we report the Binary Accuracy. This metric is computed only on the subset of data where both the human annotator and the model predicted either "Left Layout is better" or "Right Layout is better," isolating performance on the direct binary comparison.

\subsection{Results} 
As shown in Table \ref{tab:annotation_results}, our proposed DesignSense model significantly outperforms all open-source and proprietary baselines across all evaluation metrics. Our full model, \textit{DesignSense}, achieves the highest scores, with a Macro F1 of 0.456 and a Weighted F1 of 0.520. This represents a substantial improvement of 54.6\% in Macro F1 and 86.4\% in Weighted F1 over the strongest-performing proprietary baselines, GPT-5-fewshot-deepthink (0.295 Macro F1) and OpenAI-o3-fewshot (0.279 Weighted F1), respectively. An ablation study also highlights the value of our proposed perturbation method, as our full model improves upon DesignSense(w/o Perturb) by 12.3\% on Macro F1. Interestingly, while several proprietary models like OpenAI-o3-fewshot achieve a high Binary Acc (0.719), their extremely low Macro F1 (0.283) and Cohen's $\kappa$ (0.164) scores are revealing. This discrepancy suggests that while these models have some capacity to differentiate between Left Layout and Right Layout when they attempt to, they frequently fail at the overall 4-class task, likely by misclassifying preference choices as "both good" or "both bad," or vice-versa. Our model, in contrast, demonstrates strong performance in both the specific binary choice (0.732 Binary Acc) and the complete 4-class problem (0.456 Macro F1).

\begin{table}[t]
\centering
\scriptsize
\setlength{\tabcolsep}{4pt}
\begin{tabular*}{\columnwidth}{@{\extracolsep{\fill}}lccc}
\toprule
\multirow{2}{*}{\textbf{Method}} &
\multirow{2}{*}{\textbf{Judge Used for Training}} &
\multicolumn{2}{c}{\textbf{Win Rate (\%)}$\uparrow$} \\
\cmidrule(lr){3-4}
 & & GPT-4o & GPT-5 \\
\midrule
AesthetiQ \cite{patnaik2025aesthetiq} & {ViLA}\cite{10657989} & 14.27 & 15.49 \\
\midrule
AesthetiQ + AAPA (Ours) & DesignSense & \textbf{17.97} & \textbf{18.41} \\
\bottomrule
\end{tabular*}
\caption{\textbf{Effect of judge quality on AesthetiQ performance.}
Training AesthetiQ with AAPA using a stronger judge leads to higher win rates under both GPT-4o and GPT-5 evaluations, demonstrating improved alignment with aesthetic preferences.}
\label{tab:judge_quality}
\end{table}

\subsection{Layout Generator Improvements} To assess whether improving the preference model (judge) leads to better layout generation, we retrained AesthetiQ using the AAPA reinforcement learning framework \cite{patnaik2025aesthetiq}, replacing the original ViLA\cite{10657989} judge with our stronger DesignSense judge. As shown in Table \ref{tab:judge_quality}, this enhancement yields consistent gains across evaluation settings. When evaluated by GPT-4o, the win rate increased from 14.27\% to 17.97\%, while GPT-5, considered the more capable judge, reported an improvement from 15.49\% to 18.41\%, representing a relative gain of over 4 percentage points in both cases. These results clearly demonstrate that using a more reliable and fine-grained judge during AAPA training enables the layout generator to produce designs that generalize better and align more closely with human aesthetic preferences.

\subsection{Generalization to Other Distributions}
We further evaluate whether DesignSense captures general design principles beyond the Crello distribution and the AesthetiQ generator used during training. Specifically, we test (i) out-of-distribution layouts from PrismLayers (500 human-annotated pairs), and (ii) layout pairs generated by an independent model, \textbf{LayoutNUWA}, where DesignSense is applied \emph{as-is} at inference time (no additional training).

\textbf{Generalization to Out-of-Distribution Layouts.} Table \ref{tab:prism_layers} presents the evaluation on the unseen PrismLayers dataset. Despite the domain shift from our training data, DesignSense consistently outperforms state-of-the-art proprietary vision-language models. It achieves a binary accuracy of 72.2\% and a Macro F1 score of 0.41, significantly surpassing the strongest baseline, Gemini 2.5-Pro, which achieves 67.4\% binary accuracy and a 0.32 Macro F1. This performance gap confirms that DesignSense successfully internalizes fundamental spatial and aesthetic rules—such as alignment, balance, and visual hierarchy—rather than merely overfitting to the stylistic nuances of the Crello dataset.

\textbf{Generalization to Unseen Generators.} To ensure our model is generator-agnostic, we evaluate its performance on layout pairs produced by LayoutNUWA, shown in Table \ref{tab:layout_nuwa}. DesignSense demonstrates remarkable robustness, achieving an exceptional binary agreement of 92.3\%, vastly outperforming OpenAI-o3 (79.7\%) and GPT-5 (72.9\%). Most notably, in the challenging 4-class setting—which captures subjective ambiguity by requiring the model to identify "both good" or "both bad" scenarios—DesignSense reaches 68.3\% accuracy. In contrast, frontier models fail catastrophically, with none exceeding 42.5\%. These results validate DesignSense as a highly reliable, general-purpose reward model capable of evaluating layouts across diverse generation frameworks.


\begin{table*}[t]
    \centering

    \begin{minipage}{0.48\linewidth}
        \centering
        \resizebox{\linewidth}{!}{
            \begin{tabular}{lcccc}
                \toprule
                \textbf{Judge Model} & \multicolumn{4}{c}{\textbf{Metrics on PrismLayers (Unseen)}} \\
                 & Binary (\%) & 4-Class (\%) & Macro F1 & Cohen's $\kappa$ \\
                \midrule
                OpenAI-o3 & 63.9 & 42.2 & 0.26 & 0.14 \\
                Gemini 2.5-Pro & 67.4 & 45.4 & 0.32 & 0.20 \\
                GPT-5 & 61.5 & 40.2 & 0.24 & 0.11 \\
                \textbf{DesignSense (Ours)} & \textbf{72.2} & \textbf{46.4} & \textbf{0.41} & \textbf{0.24} \\
                \bottomrule
            \end{tabular}
        }
        \caption{Generalization to Out-of-Distribution Data: DesignSense maintains high agreement on PrismLayers (unseen).}
        \label{tab:prism_layers}
    \end{minipage}
    \hfill 
    \begin{minipage}{0.48\linewidth}
        \centering
        \resizebox{\linewidth}{!}{
            \begin{tabular}{lcccc}
                \toprule
                \textbf{Judge Model} & \multicolumn{4}{c}{\textbf{Metrics on LayoutNUWA Pairs}} \\
                 & Binary (\%) & 4-Class (\%) & Macro F1 & Cohen's $\kappa$ \\
                \midrule
                OpenAI-o3 & 79.7 & 28.0 & 0.21 & 0.12 \\
                Gemini 2.5-Pro & 75.6 & 42.5 & 0.31 & 0.20 \\
                GPT-5 & 72.9 & 27.3 & 0.21 & 0.11 \\
                \textbf{DesignSense (Ours)} & \textbf{92.3} & \textbf{68.3} & \textbf{0.37} & \textbf{0.25} \\
                \bottomrule
            \end{tabular}
        }
        \caption{Generalization to Other Generators: DesignSense maintains high agreement metrics on LayoutNUWA generated layouts.}
        \label{tab:layout_nuwa}
    \end{minipage}
\end{table*}

\subsection{Inference time scaling} While we have a trained DesignSense judge, we aim to evaluate whether it can support inference-time scaling, allowing the model to produce layouts that better align with human preferences through additional computation. To this end, we use the state-of-the-art layout generation model AesthetiQ~\cite{patnaik2025aesthetiq} and generate 10 layouts per sample instead of the default single output. We then apply DesignSense judge to perform pairwise preference predictions across all candidates and rank them to select the best layout. Finally, to assess whether inference-time scaling with DesignSense improves human preference alignment, we use GPT-5 as the judge and compare the original AesthetiQ outputs with the scaled results.

Evaluating this strategy on 20\% of the Crello test set used in AesthetiQ~\cite{patnaik2025aesthetiq}, we observed an improvement in the GPT-5 win rate over the ground-truth layout from 16.6\% to 20.2\%, representing a gain of nearly four points. This demonstrates that DesignSense effectively captures human preferences and improves layout selection through inference-time scaling.

\section{Conclusion}

We introduce \OURNAME, a large-scale dataset of 10,235 graphic layout pairs with human preference annotations, designed to advance the training of reward models for layout generation tasks. Our three-fold contributions span generation, annotation and judgement of graphic design layouts. Starting from a multi-stage relayout pipeline that allows annotators to mark preferences on a high-quality dataset, we capture human preferences across diverse aspect ratios. Further, the annotation methodology provides a clear disambiguation option which is not tackled by most current open source and proprietary VLMs. Finally, our judge provides a strong reward signal for aligning downstream layout generation with human preferences. Thus, our proposed methodology provides a complete and holistic framework.

\bibliographystyle{splncs04}
\bibliography{main}
\end{document}